\documentclass{article}
\usepackage{spconf,amsmath,graphicx}
\usepackage{subcaption} 

\usepackage{graphicx,verbatim}
\usepackage{threeparttable}
\usepackage{booktabs}
\usepackage{makecell}
\usepackage{multirow}
\usepackage{amssymb}
\usepackage{arydshln}
\usepackage{marvosym}

\usepackage{enumitem}
\setlist{nosep, leftmargin=14pt}
\usepackage{url}

\usepackage{mwe} 

\title{Cross-Domain Vessel Segmentation via Latent Similarity Mining and Iterative Co-Optimization}
\newsavebox\CBox
\def\textBF#1{\sbox\CBox{#1}\resizebox{\wd\CBox}{\ht\CBox}{\textbf{#1}}}

\name{Zhanqiang Guo$^{1,2}$, Jianjiang Feng$^{1,2}\textsuperscript{(\Letter)}$, and Jie Zhou$^{1,2}$}
\address{$^{1}$ Department of Automation, Tsinghua University, Beijing, China \\
	$^{2}$ Beijing National Research Center for Information Science and Technology, Beijing, China}

\begin{document}
%
\maketitle
\begin{abstract}
Retinal vessel segmentation serves as a critical prerequisite for automated diagnosis of retinal pathologies. While recent advances in Convolutional Neural Networks (CNNs) have demonstrated promising performance in this task, significant performance degradation occurs when domain shifts exist between training and testing data. To address these limitations, we propose a novel domain transfer framework that leverages latent vascular similarity across domains and iterative co-optimization of generation and segmentation networks. Specifically, we first pre-train generation networks for source and target domains. Subsequently, the pretrained source-domain conditional diffusion model performs deterministic inversion to establish intermediate latent representations of vascular images, creating domain-agnostic prototypes for target synthesis. Finally, we develop an iterative refinement strategy where segmentation network and generative model undergo mutual optimization through cyclic parameter updating. This co-evolution process enables simultaneous enhancement of cross-domain image synthesis quality and segmentation accuracy. Experiments demonstrate that our framework achieves state-of-the-art performance in cross-domain retinal vessel segmentation, particularly in challenging clinical scenarios with significant modality discrepancies.
\end{abstract}

\section{Introduction}
\label{sec:intro}

\begin{figure}[tbp]
    \centering
    \begin{subfigure}[b]{0.07\textwidth}
        \centering
        \includegraphics[width=\textwidth]{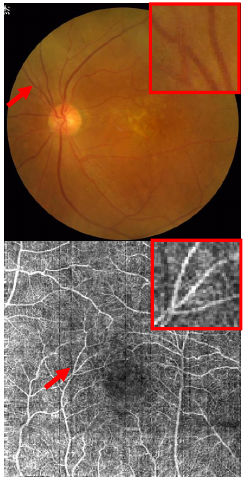}
        \caption{}
        \label{fig:sub1}
    \end{subfigure}
    \hfill
    \begin{subfigure}[b]{0.19\textwidth}
        \centering
        \includegraphics[width=\textwidth]{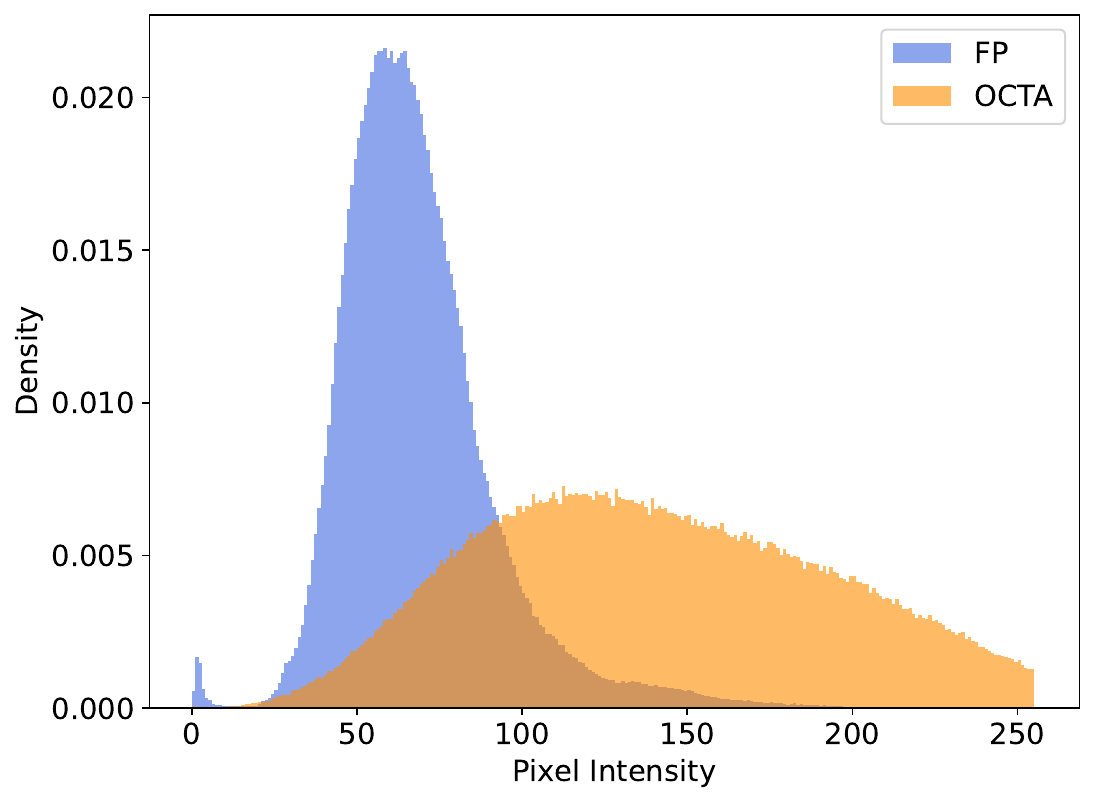}
        \caption{}
        \label{fig:sub2}
    \end{subfigure}
    \hfill
    \begin{subfigure}[b]{0.21\textwidth}
        \centering
        \includegraphics[width=\textwidth]{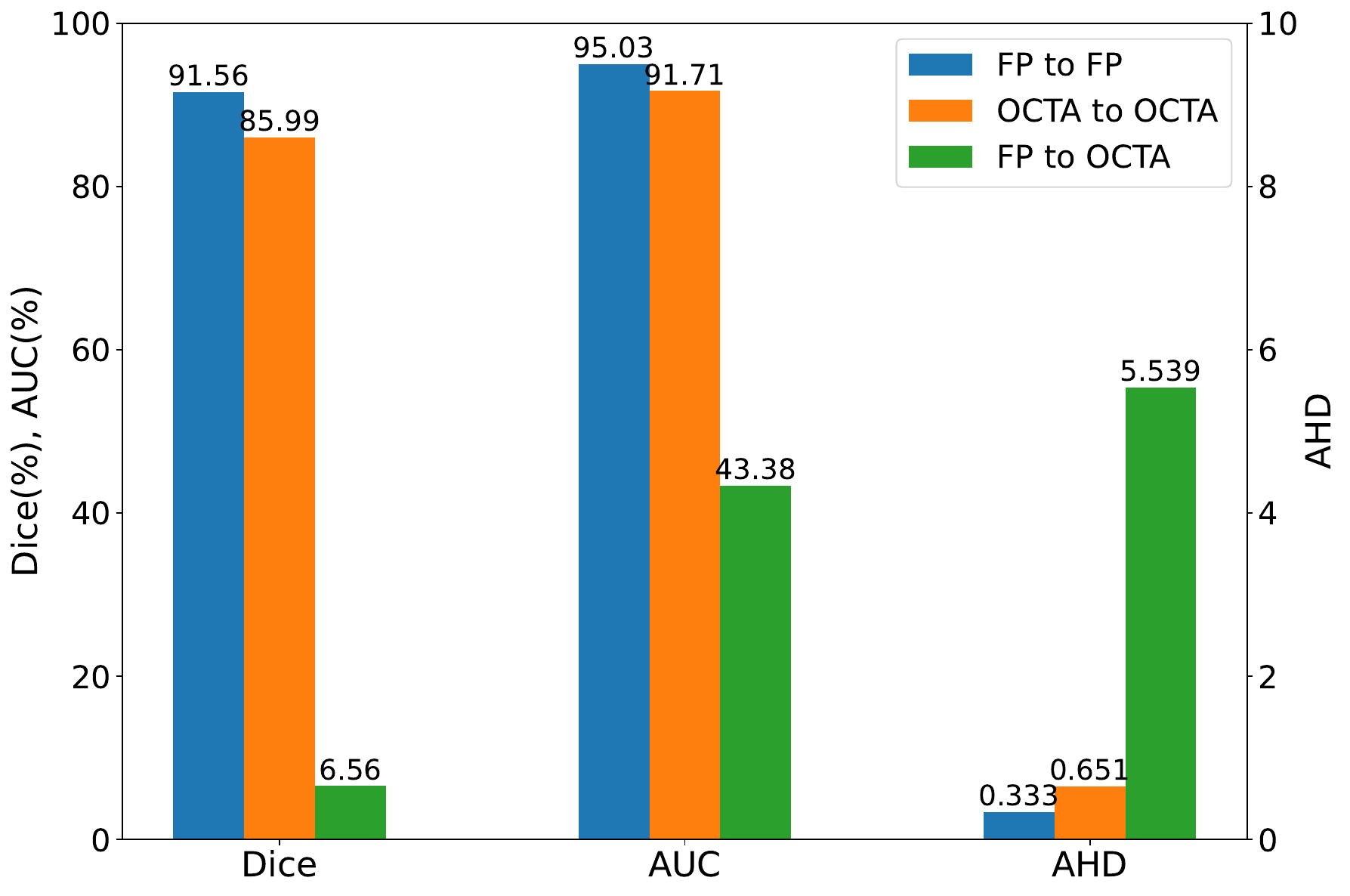}
        \caption{}
        \label{fig:sub3}
    \end{subfigure}
    \vspace{-.18cm}
    \caption{Cross-modality discrepancy visualization. (a): Samples from FP and OCTA modalities. (b): Intensity distribution histograms. (c): Segmentation performance of Unet~\cite{ronneberger2015u} under in-domain and cross-domain scenarios.} 
    \label{fig:three_subfigures}
    \vspace{-.5cm}
\end{figure}

Accurate segmentation of retinal vasculature serves as a critical prerequisite for automated detection of ocular pathologies.
Recent advances in Convolutional Neural Networks (CNNs) have demonstrated remarkable performance in retinal vessel segmentation when training and testing data share similar distributions~\cite{zhou2025masked}. However, significant performance degradation occurs when deploying these models across different imaging modalities, such as between Fundus Photography (FP) and Optical Coherence Tomography Angiography (OCTA) images, due to their substantial intensity distribution discrepancies (as illustrated in Fig.~\ref{fig:sub1}-\ref{fig:sub2}). This cross-modality challenge leads to dramatic segmentation quality deterioration, as evidenced in Fig.~\ref{fig:sub3}.

Unsupervised domain adaptation (UDA) has emerged as a promising solution to bridge this domain shift by transferring knowledge from label-rich source domain to unlabeled target domain. Pseudo-label-based methods, an UDA paradigm, typically generate initial target domain annotations using source-trained networks, followed by refinement strategies such as uncertainty estimation~\cite{wu2024fpl}, teacher-student frameworks~\cite{hu2024chebyshev}, or contrastive learning~\cite{zhang2025link}. Nevertheless, these approaches exhibit strong dependency on initial pseudo-label quality, which becomes particularly problematic when confronting substantial domain shifts between modalities. Alternative UDA methodologies employ Generative Adversarial Networks (GANs) to minimize inter-domain discrepancies through either image-level style transfer~\cite{xu2023novel} or feature-space alignment~\cite{han2021deep}. While effective in certain scenarios, these approaches primarily focus on global domain adaptation and often overlook critical local structural details and shared semantic information between domains~\cite{xie2023sepico}. This limitation becomes especially pronounced in retinal image analysis where fine vascular structures and pathological features require attention to details.

Despite significant appearance discrepancies across imaging modalities, we observe that local vascular structures maintain consistent morphological characteristics, such as their elongated tubular patterns (as demonstrated in the zoomed-in regions of Fig.~\ref{fig:sub1}). 
Leveraging the inherent similarities in semantic information is crucial for establishing effective cross-modality bridges,
especially when dealing with the sparsity and complexity of vascular networks. 
The recent success of diffusion models~\cite{ho2020denoising,nichol2021improved} in generative tasks has inspired new approaches for latent feature representation. Specifically, Denoising Diffusion Implicit Models (DDIM)~\cite{song2020denoising} enable deterministic inversion techniques that can retrieve latent codes while preserving essential semantic information~\cite{xie2024diffdgss,cho2024one}. Xie et al.~\cite{xie2024diffdgss} demonstrated that DDIM-based deterministic noise injection could generate generalized vascular representations for cross-domain segmentation. However, the inherent noise in latent space may compromise segmentation accuracy. 

Inspired by these advancements, we propose a novel framework for unsupervised vascular domain adaptation that jointly leverages latent feature similarity and iterative co-optimization. Specifically, we first train a conditional Denoising Diffusion Probabilistic Model (DDPM) generator on source domain images, employing DDIM-based deterministic inversion to obtain latent representations. The preservation strength of vascular structures is controlled through time-step encoding and vascular semantic guidance. Subsequently, the latent vascular representations serve as intermediate inputs to a pre-trained target domain generator, producing synthetic target domain images. These generated images, paired with source domain annotations, are then used to iteratively optimize the target domain segmentation network as well as the generation network conditioned on segmentation results.
This co-optimization strategy simultaneously enhances both cross-domain image synthesis quality and segmentation accuracy. 
Extensive experiments demonstrate that our method outperforms state-of-the-art approaches, particularly in preserving fine vascular structures.

\vspace{-.2cm}

\section{Method}

\begin{figure}[h]
	\centering
	\centerline{\includegraphics[width=0.5\textwidth]{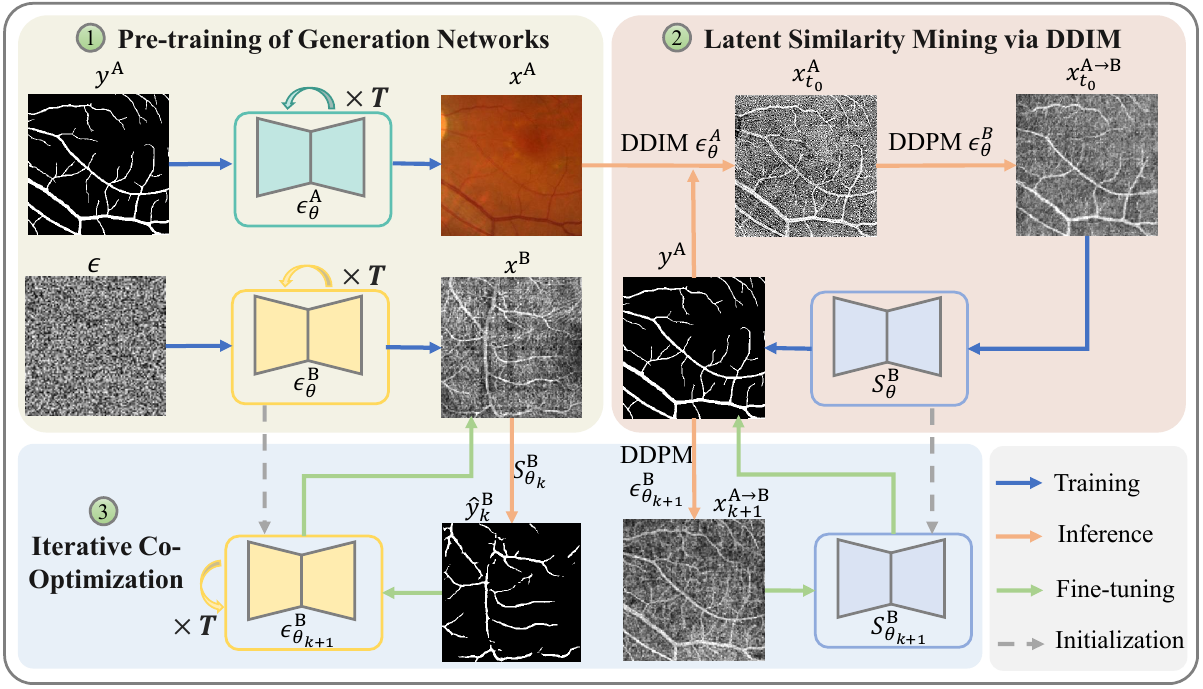}}
	\caption{Overview of the proposed framework.}
	\label{fig:framework}
    \vspace{-0.1cm}
\end{figure}

Let $D_\mathrm{A}=\{(x_i^\mathrm{A}, y_i^\mathrm{A})\}_{i=1}^{M}$ denotes the source domain dataset with corresponding annotations, and $D_\mathrm{B}=\{(x_i^\mathrm{B})\}_{i=1}^{N}$ represents the unlabeled target domain dataset. Our objective is to leverage the labeled source domain data to achieve accurate segmentation on the target domain. As illustrated in Fig.~\ref{fig:framework}, our framework comprises three key stages: 1) Pre-training of vessel image generation networks (Sec.~\ref{subsec:ddpm}), 2) Vascular latent similarity mining and target domain image generation (Sec.~\ref{subsec:ddim}), and 3) Iterative co-optimization of segmentation network and conditional generation network (Sec.~\ref{subsec:iter}). 
\vspace{-.2cm}

\subsection{DDPM and Pre-training}
\label{subsec:ddpm}

DDPMs are latent generative models that operate through forward and reverse diffusion processes. The forward process gradually corrupts training data $x_0\sim q(x_0)$ by iteratively adding Gaussian noise according to a predefined variance schedule $\{\beta_i\}_{t=0}^T$. This Markov process is formally defined as $q\left(x_{0: T}\right)=q\left(x_0\right) \prod_{t=1}^T q\left(x_t \mid x_{t-1}\right)$, where
\vspace{-.1cm}
\begin{equation}
    q\left(x_t \mid x_{t-1}\right)=\mathcal{N}\left(\sqrt{1-\beta_t} x_{t-1}, \beta_t \mathbf{I}\right).
\end{equation}
\vspace{-.1cm}
The reverse process generates new samples by learning to reconstruct the original data through a parameterized model that approximates the posterior distribution at each diffusion step: $p_\theta\left(x_{t-1} \mid x_t\right)=\mathcal{N}\left(\mu_\theta\left(x_t, t \right), \sigma_t^2 \mathbf{I}\right)$, with
\begin{equation}
\label{eq:posterior}
    \mu_\theta\left(x_t, t\right)=\frac{1}{\sqrt{1-\beta_t}}\left(x_t-\frac{\beta_t}{\sqrt{1-\alpha_t}} \epsilon_\theta\left(x_t, t\right)\right),
\end{equation}
where $\epsilon_\theta$ represents a learnable network that predicts the reverse diffusion steps. The training objective follows variational inference principles, minimizing the variational lower bound of the negative log-likelihood:
\vspace{-.1cm}
\begin{equation}
\label{eq:loss}
    \mathcal{L}_t=\mathbb{E}_{x_0\sim q(x_0)}\mathbb{E}_{\epsilon_t\sim \mathcal{N}(0, \mathbf{I})}\left\|\epsilon_t-\epsilon_\theta\left(x_t, t\right)\right\|^2.
\end{equation}
\vspace{-.0cm}
In the first stage, we pre-train two networks: 1) A conditional generation network on source domain $D_\mathrm{A}$, where $\epsilon_\theta$ in Eqs.~\ref{eq:posterior} and \ref{eq:loss} is parameterized as $\epsilon_\theta^\mathrm{A}(x_t^\mathrm{A},t,y^\mathrm{A})$ to incorporate conditional information. 2) An unconditional generation network on target domain $D_\mathrm{B}$, with $\epsilon_\theta$ implemented as $\epsilon_\theta^\mathrm{B}(x_t^\mathrm{B},t)$.

\subsection{Latent Similarity Mining via DDIM}
\label{subsec:ddim}

Building upon DDPM, \cite{song2020denoising} proposed a non-Markovian forward process and derived its corresponding reverse process:
\begin{equation}
    x_{t-1}=\sqrt{\alpha_{t-1}} f_\theta\left(x_t, t\right)+\sqrt{1-\alpha_{t-1}} \epsilon_\theta\left(x_t, t\right),
\end{equation}
\begin{equation}
f_\theta\left(x_t, t\right)=\frac{x_t-\sqrt{1-\alpha_t} \epsilon_\theta\left(x_t, t\right)}{\sqrt{\alpha_t}}.
\end{equation}
Furthermore, DDIM introduces an image encoding method by deriving an ordinary differential equation (ODE) corresponding to the reverse process:
\begin{equation}
    x_{t+1}=\sqrt{\alpha_{t+1}} f_\theta\left(x_t, t\right)+\sqrt{1-\alpha_{t+1}} \epsilon_\theta\left(x_t, t\right),
\end{equation}
which, known as the deterministic reverse DDIM process~\cite{song2020denoising,dhariwal2021diffusion}, allows images to be noised through the trained $\epsilon_\theta$ while preserving essential image information.

Inspired by~\cite{cho2024one,preechakul2022diffusion}, the latent space images obtained through the deterministic reverse DDIM process in conditional DDPM networks inherently encode input conditional information. Leveraging the pre-trained source domain conditional generation network $\epsilon_\theta^{\mathrm{A}}$, we apply $t_0$ steps of deterministic noising to source domain images $x^\mathrm{A}$, yielding structural latent codes $x_{t_0}^\mathrm{A}$: $x_{t+1}^{\mathrm{A}}=\sqrt{\alpha_{t+1}} f_\theta^{\mathrm{A}}\left(x_t^{\mathrm{A}}, t, y^\mathrm{A}\right)+\sqrt{1-\alpha_{t+1}} \epsilon_\theta^{\mathrm{A}}\left(x_t^{\mathrm{A}}, t, y^\mathrm{A}\right), t={0,1,2...t_0-1}$.

The resulting latent codes are domain-independent while preserving vascular structures due to the guidance of annotations $y^\mathrm{A}$. By utilizing the latent codes as intermediate inputs to the target domain generation network, we can generate corresponding target domain images $x_{t_0}^{\mathrm{A}\rightarrow \mathrm{B}}$ through the reverse denoising process using the pre-trained $\epsilon_\theta^{\mathrm{B}}$. The synthesized target domain images $x_{t_0}^{\mathrm{A}\rightarrow \mathrm{B}}$, paired with their corresponding source domain labels $y^{\mathrm{A}}$, form a training set for the target domain segmentation network $S_\theta^{\mathrm{B}}$.

\subsection{Iterative Co-Optimization}
\label{subsec:iter}

Due to significant cross-modality discrepancies, the initial target domain images generated through DDIM inversion followed by DDPM backward process ($x_{t_0}^{\mathrm{A}\rightarrow \mathrm{B}}$) exhibit still some distributional shifts from real target domain images ($x^\mathrm{B}$), as illustrated in Fig.~\ref{fig:framework}. This discrepancy may lead to suboptimal performance of the segmentation network. To address this limitation, we propose an iterative co-optimization strategy that jointly refines both the target domain conditional generation network and segmentation network. 

We regard the pre-trained generative network $\epsilon_\theta^{\mathrm{B}}$ and segmentation network $S_\theta^{\mathrm{B}}$, trained as described in Sec.~\ref{subsec:ddim}, as the initial models, denoted as $\epsilon_{\theta_0}^{\mathrm{B}}$ and $S_{\theta_0}^{\mathrm{B}}$, respectively. At the $k$-th iteration, the generation and segmentation networks are represented as $\epsilon_{\theta_{k}}^{\mathrm{B}}$ and $S_{\theta_k}^{\mathrm{B}}$. 
The target domain images $x^\mathrm{B}$ from dataset $D_\mathrm{B}$ are fed into segmentation network $S_{\theta_k}^{\mathrm{B}}$ to obtain the corresponding segmentation results $\hat{y}_{k}^\mathrm{B}$. 
The paired data $(x^\mathrm{B},\hat{y}_{k}^\mathrm{B})$ are used to train the conditional generation network $\epsilon_{\theta_{k+1}}^{\mathrm{B}}$, initialized with $\epsilon_{\theta_k}^{\mathrm{B}}$ parameters. 
Subsequently, conditioned on the source domain annotations $y^\mathrm{A}$, the trained $\epsilon_{\theta_{k+1}}^{\mathrm{B}}$ generates the corresponding target domain images $x_{k+1}^{\mathrm{A}\rightarrow \mathrm{B}}$, which are employed to further refine the segmentation network, resulting in $S_{\theta_{k+1}}^{\mathrm{B}}$ for the next iteration.

The segmentation network is optimized using a combination of Dice loss and Cross-Entropy loss. For the generation network, we enhance Eq. \ref{eq:loss} by incorporating additional supervision on noise prediction in vascular regions, thereby improving the generation of fine vascular structures.

\section{Experiments and Results}

\subsection{Experimental Setup} 

\subsubsection{Dataset and Evaluation Metrics.}

Following the previous researches~\cite{xie2024diffdgss,lyu2022aadg}, we employ FP as the source domain and OCTA as the target domain. 
The source domain data is derived from the FIVES dataset~\cite{jin2022fives}, from which we select 563 annotated samples to construct $D_\mathrm{A}$.
For the target domain, we adopt two publicly available OCTA datasets: OCTA-500~\cite{li2020ipn} and ROSE~\cite{ma2020rose}. From these, we select 450 and 30 samples, respectively, to form $D_\mathrm{B}$. To evaluate performance, we employ four standard metrics: Dice Similarity Coefficient (DSC), Area Under the Curve (AUC), Accuracy (ACC), and Average Hausdorff Distance (AHD). 

\subsubsection{Implementation Details.}
Our framework is implemented on the PyTorch platform, utilizing two NVIDIA GeForce GTX 4090 GPUs. The generation network architecture follows~\cite{nichol2021improved}, with $T=1000$ diffusion steps. For DDIM inversion, we set $t_0=15$ and use 50 DDIM steps. During training, images are randomly cropped into $256\times 256$ patches before feeding into the network. The AdamW optimizer is employed for generation training, with the initial learning rate of $1\times \mathrm{e}^{-4}$. For the segmentation network, we use the Adam optimizer with a learning rate of $1\times \mathrm{e}^{-3}$. Our code is available at: \url{https://github.com/gzq17/Diffusion-UDA}. 

\begin{table}[b]
	\centering
	\scriptsize
    \renewcommand\arraystretch{0.95}
	\caption{Comparison with other methods on two datasets.}
	\label{tab1}
    \begin{tabular}{ >{\centering\arraybackslash}m{1.0cm} : >{\centering\arraybackslash}m{1.3cm} : >{\centering\arraybackslash}m{0.9cm}  >{\centering\arraybackslash}m{0.9cm}  >{\centering\arraybackslash}m{0.9cm}  >{\centering\arraybackslash}m{0.8cm} }
		\hline
        \textbf{Data} &\textbf{Method} & \textbf{DSC(\%)} & \textbf{AUC(\%)} & \textbf{ACC(\%)} & \textbf{AHD}   \\
		\hline
        \multirow{7}*[-3pt]{\textbf{OCTA-500}} 
		  & Baseline~\cite{ronneberger2015u}       &  6.56  & 43.38 & 68.40  & 5.539  \\
        & AADG~\cite{lyu2022aadg}           &  50.78 &   -   &   -    &   -   \\
        & DGSS~\cite{xie2024diffdgss}       &  61.99 &   -   &   -    &   -   \\
        & AADG$^*$~\cite{lyu2022aadg}       &  57.33 & 74.87 & 93.96  & 1.853 \\
        & SFDA~\cite{hu2024chebyshev}           &  20.65 & 56.23 & 91.43  & 5.623  \\
        & Ours (Iter-0)  &  69.02 & 85.95 & 94.60  & 2.301 \\
        & Ours (Final)   &  \textBF{78.33} & \textBF{89.96} & \textBF{96.32}  & \textBF{1.001} \\
		\hline
        \multirow{7}*[-3pt]{\textbf{ROSE}} 
        & Baseline~\cite{ronneberger2015u}  & 11.89 & 49.41 & 76.95 & 5.607 \\
        & AADG~\cite{lyu2022aadg}           & 61.57 &   -   &   -   &   -   \\
        & DGSS~\cite{xie2024diffdgss}       & 66.88 &   -   &   -   &   -   \\
        & AADG$^*$~\cite{lyu2022aadg}       & 67.75 & 79.32 & 91.63 & 1.506 \\
        & SFDA~\cite{hu2024chebyshev}       & 52.50 & 69.81 & 88.88 & 2.128 \\
        & Ours (Iter-0)  & 64.27 & 75.65 & 91.61 & 2.818 \\
        & Ours (Final)   &  \textBF{75.28} & \textBF{86.05} & \textBF{92.77}  & \textBF{1.317} \\
        \hline
	\end{tabular}
\end{table}

\begin{figure*}[t]
	\centering
	\centerline{\includegraphics[width=0.75\textwidth]{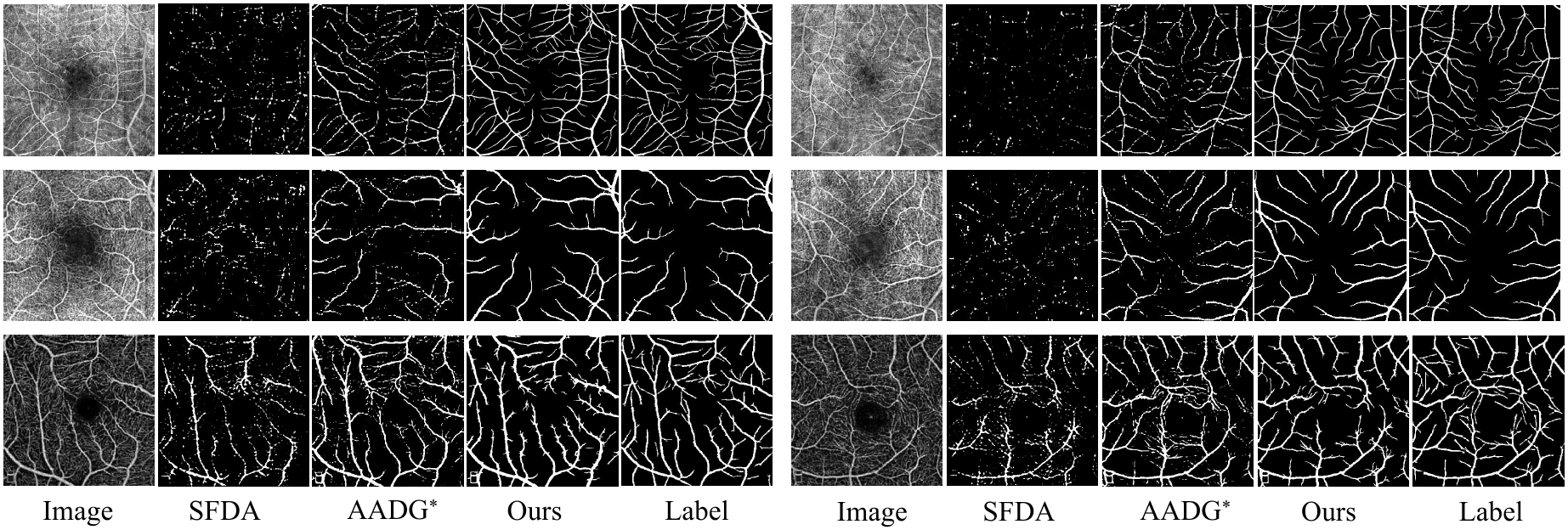}}
	\caption{Illustration of results. The first two rows are the results from the OCTA-500, while the third row is from the ROSE.}
	\label{fig:seg}
    \vspace{-.4cm}
\end{figure*}

\subsection{Results and Comparison}

To evaluate the effectiveness of our proposed framework, we conduct comprehensive comparisons with several UDA methods. Notably, AADG$^*$ represents our re-implementation of~\cite{lyu2022aadg}, augmented with the FIVES dataset to ensure fair comparison. The performance of the source-trained model, when directly applied to segment images from the target domain, is regarded as the Baseline.

As shown in Table~\ref{tab1}, the baseline model exhibits inadequate performance, highlighting the substantial distributional shift between source and target domains.
We report results for two variants of the proposed method: Ours (Iter-0) represents the initial segmentation performance without co-optimization, while Ours (Final) denotes the performance after iterative refinement. 
Notably, the substantial improvements in DSC scores—from (Iter-0) to (Final)—of 9.31\% and 11.01\% across the two datasets, respectively, substantiate the effectiveness of our iterative optimization strategy.
Furthermore, our proposed method outperforms the compared methods across all evaluation metrics, indicating the effectiveness and robustness of our approach in challenging scenarios. The strong performance of (Iter-0) provides compelling evidence that DDIM inversion effectively captures latent structural similarities across modalities.

Fig.~\ref{fig:seg} presents qualitative comparisons of segmentation results on two different datasets. SFDA~\cite{hu2024chebyshev} exhibits suboptimal performance, likely due to its heavy reliance on initial segmentation quality and sensitivity to substantial cross-modality discrepancies. By leveraging latent structural similarities of vascular patterns across modalities, our framework effectively transfers knowledge from source domain labels to target domain images. This enables accurate vessel boundary detection and maintains superior vascular connectivity in the segmentation results, as evidenced by the visual comparisons. 
\vspace{-.2cm}
\begin{figure}[h]
    \centering
    \begin{subfigure}[b]{0.2\textwidth}
        \centering
        \includegraphics[width=\textwidth]{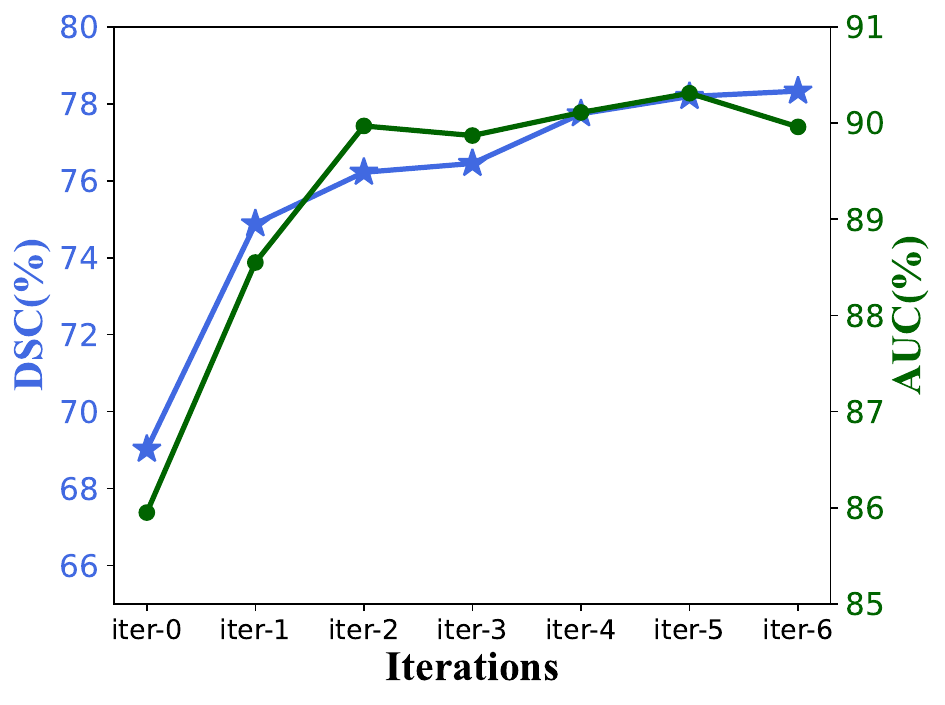}
        \caption{DSC $\&$ AUC vs. Iter}
        \label{fig:dsc}
    \end{subfigure}
    \hfill
    \begin{subfigure}[b]{0.2\textwidth}
        \centering
        \includegraphics[width=\textwidth]{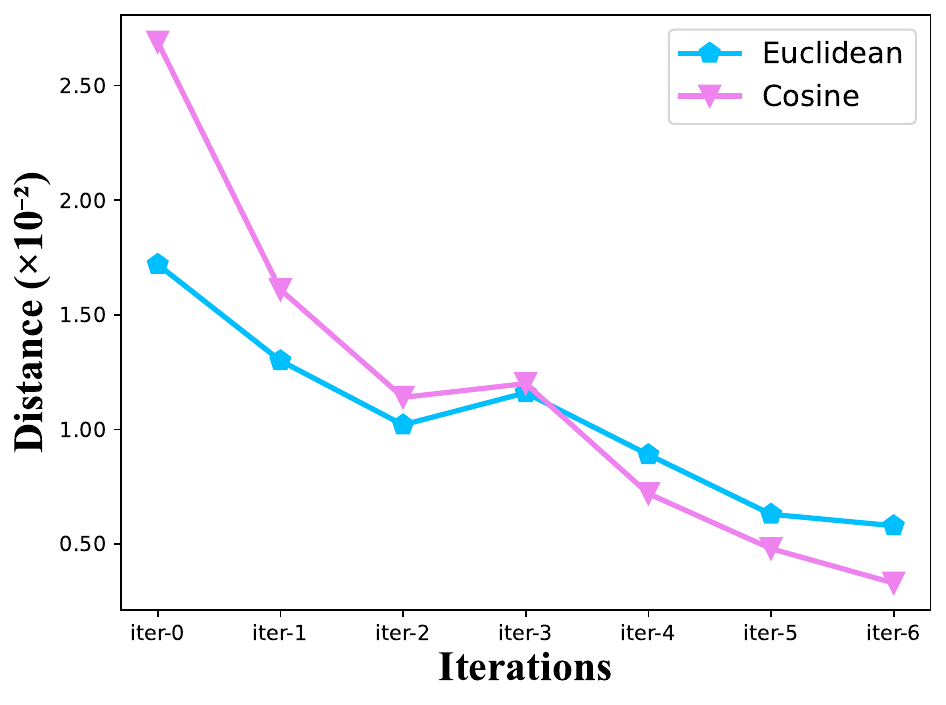}
        \caption{Distance vs. Iter}
        \label{fig:dis}
    \end{subfigure}
    \vspace{-.18cm}
    \caption{Performance evolution during iterative optimization.}
    \label{fig:Iterations}
    \vspace{-.4cm}
\end{figure}
\subsection{Iteration Process and Latent Similarity Mining}

\textbf{Iteration Process.} To further demonstrate the effectiveness of our iterative co-optimization strategy, we present the performance evolution across different iteration steps on the OCTA-500 dataset (as shown in Fig.~\ref{fig:Iterations}). The results reveal a consistent performance improvement across all evaluation metrics as the iteration progresses. This trend indicates that the quality of segmentation results used as input to the conditional generation network progressively improves (Fig.~\ref{fig:dsc}), thereby enhancing the network's capability to recover target domain image details from the source domain labels. Consequently, the discrepancy between generated target domain images and real images diminishes (Fig.~\ref{fig:dis}), leading to continuous improvement in the segmentation network's performance. Notably, the most significant performance gain occurs during the first iteration. This can be attributed to the fact that the initial generation results of (Iter-0) are obtained through unconditional generation from latent codes of source domain images, which may still exhibit differences from real target domain data due to limitations in DDIM inversion. The performance stabilizes after six iterations, leading us to select $K=6$ as our final model (Ours (Final)).

\textbf{Latent Similarity Mining.} The effectiveness of our DDIM-based latent similarity mining is quantitatively demonstrated by the significant reduction in distributional distances. Specifically, the Euclidean and Cosine distances of image intensity distributions (between the target domain images) decrease from 0.102 and 0.513 to 0.017 and 0.027, respectively, when comparing source domain images with the initial generation results from (Iter-0). This reduction indicates that our method effectively leverages the inherent semantic similarities in vascular structures across domains. Consequently, the initial segmentation results achieve a remarkable 62.46\% improvement in DSC compared to the baseline, demonstrating the effectiveness of our approach in bridging the domain gap even before iterative refinement. 

\section{Conclusion}

We propose an unsupervised domain adaptation framework to address cross-modality discrepancies in retinal vessel imaging through latent similarity mining and iterative co-optimization. Using DDIM deterministic inversion, source-domain images are encoded into latent representations guided by vessel annotations to preserve vascular semantics, which then facilitate target-domain image synthesis. An iterative co-optimization strategy jointly refines the generation and segmentation networks, where segmentation results condition the generator, and generated images enhance segmentation training. Experiments confirm the framework’s effectiveness in narrowing modality gaps and improving cross-domain vessel segmentation. Future work will extend the method to broader vascular datasets and pursue fully end-to-end training.

\bibliographystyle{IEEEbib}
\bibliography{strings,refs}

@article{zhou2025masked,
  title={Masked vascular structure segmentation and completion in retinal images},
  author={Zhou, Yi and Ahmed, Thiara Sana and Wang, Meng and Newman, Eric A and Schmetterer, Leopold and Fu, Huazhu and Cheng, Jun and Tan, Bingyao},
  journal={IEEE Transactions on Medical Imaging},
  year={2025},
  publisher={IEEE}
}

@inproceedings{zhang2025link,
  title={Link-based Contrastive Learning for One-Shot Unsupervised Domain Adaptation},
  author={Zhang, Yue and Bin, Mingyue and Zhang, Yuyang and Wang, Zhongyuan and Han, Zhen and Liang, Chao},
  booktitle={Proceedings of the Computer Vision and Pattern Recognition Conference},
  pages={4916--4926},
  year={2025}
}

@article{xu2023novel,
  title={A novel one-to-multiple unsupervised domain adaptation framework for abdominal organ segmentation},
  author={Xu, Xiaowei and Chen, Yinan and Wu, Jianghao and Lu, Jiangshan and Ye, Yuxiang and Huang, Yechong and Dou, Xin and Li, Kang and Wang, Guotai and Zhang, Shaoting and others},
  journal={Medical Image Analysis},
  volume={88},
  pages={102873},
  year={2023},
  publisher={Elsevier}
}

@article{han2021deep,
  title={Deep symmetric adaptation network for cross-modality medical image segmentation},
  author={Han, Xiaoting and Qi, Lei and Yu, Qian and Zhou, Ziqi and Zheng, Yefeng and Shi, Yinghuan and Gao, Yang},
  journal={IEEE Transactions on Medical Imaging},
  volume={41},
  number={1},
  pages={121--132},
  year={2021},
  publisher={IEEE}
}

@article{xie2023sepico,
  title={Sepico: Semantic-guided pixel contrast for domain adaptive semantic segmentation},
  author={Xie, Binhui and Li, Shuang and Li, Mingjia and Liu, Chi Harold and Huang, Gao and Wang, Guoren},
  journal={IEEE Transactions on Pattern Analysis and Machine Intelligence},
  volume={45},
  number={7},
  pages={9004--9021},
  year={2023},
  publisher={IEEE}
}

@article{wu2024fpl,
  title={FPL+: Filtered pseudo label-based unsupervised cross-modality adaptation for {3D} medical image segmentation},
  author={Wu, Jianghao and Guo, Dong and Wang, Guotai and Yue, Qiang and Yu, Huijun and Li, Kang and Zhang, Shaoting},
  journal={IEEE Transactions on Medical Imaging},
  year={2024},
  volume={43},
  number={9},
  pages={3098-3109},
  publisher={IEEE}
}

@article{hu2024chebyshev,
  title={A Chebyshev Confidence Guided Source-Free Domain Adaptation Framework for Medical Image Segmentation},
  author={Hu, Jiesi and Yang, Yanwu and Guo, Xutao and Wang, Jinghua and Ma, Ting},
  journal={IEEE Journal of Biomedical and Health Informatics},
  year={2024},
  volume={28},
  number={9},
  pages={5473-5486},
  publisher={IEEE}
}

@article{ho2020denoising,
  title={Denoising diffusion probabilistic models},
  author={Ho, Jonathan and Jain, Ajay and Abbeel, Pieter},
  journal={Advances in Neural Information Processing Systems},
  volume={33},
  pages={6840--6851},
  year={2020}
}

@inproceedings{nichol2021improved,
  title={Improved denoising diffusion probabilistic models},
  author={Nichol, Alexander Quinn and Dhariwal, Prafulla},
  booktitle={International Conference on Machine Learning},
  pages={8162--8171},
  year={2021},
  organization={PMLR}
}

@article{song2020denoising,
  title={Denoising diffusion implicit models},
  author={Song, Jiaming and Meng, Chenlin and Ermon, Stefano},
  journal={arXiv preprint arXiv:2010.02502},
  year={2020}
}

@inproceedings{xie2024diffdgss,
  title={{DiffDGSS}: Generalizable Retinal Image Segmentation with Deterministic Representation from Diffusion Models},
  author={Xie, Yingpeng and Qu, Junlong and Xie, Hai and Wang, Tianfu and Lei, Baiying},
  booktitle={International Conference on Medical Image Computing and Computer-Assisted Intervention},
  pages={166--176},
  year={2024},
  organization={Springer}
}

@inproceedings{cho2024one,
  title={One-shot structure-aware stylized image synthesis},
  author={Cho, Hansam and Lee, Jonghyun and Chang, Seunggyu and Jeong, Yonghyun},
  booktitle={Proceedings of the IEEE/CVF Conference on Computer Vision and Pattern Recognition},
  pages={8302--8311},
  year={2024}
}

@article{dhariwal2021diffusion,
  title={Diffusion models beat {GANs} on image synthesis},
  author={Dhariwal, Prafulla and Nichol, Alexander},
  journal={Advances in Neural Information Processing Systems},
  volume={34},
  pages={8780--8794},
  year={2021}
}

@inproceedings{preechakul2022diffusion,
  title={Diffusion autoencoders: Toward a meaningful and decodable representation},
  author={Preechakul, Konpat and Chatthee, Nattanat and Wizadwongsa, Suttisak and Suwajanakorn, Supasorn},
  booktitle={Proceedings of the IEEE/CVF Conference on Computer Vision and Pattern Recognition},
  pages={10619--10629},
  year={2022}
}

@article{lyu2022aadg,
  title={{AADG}: automatic augmentation for domain generalization on retinal image segmentation},
  author={Lyu, Junyan and Zhang, Yiqi and Huang, Yijin and Lin, Li and Cheng, Pujin and Tang, Xiaoying},
  journal={IEEE Transactions on Medical Imaging},
  volume={41},
  number={12},
  pages={3699--3711},
  year={2022},
  publisher={IEEE}
}

@article{jin2022fives,
  title={Fives: A fundus image dataset for artificial Intelligence based vessel segmentation},
  author={Jin, Kai and Huang, Xingru and Zhou, Jingxing and Li, Yunxiang and Yan, Yan and Sun, Yibao and Zhang, Qianni and Wang, Yaqi and Ye, Juan},
  journal={Scientific Data},
  volume={9},
  number={1},
  pages={475},
  year={2022},
  publisher={Nature Publishing Group UK London}
}

@article{li2020ipn,
  title={Ipn-v2 and {OCTA}-500: Methodology and dataset for retinal image segmentation},
  author={Li, Mingchao and Zhang, Yuhan and Ji, Zexuan and Xie, Keren and Yuan, Songtao and Liu, Qinghuai and Chen, Qiang},
  journal={arXiv preprint arXiv:2012.07261},
  volume={5},
  number={6},
  pages={7},
  year={2020}
}

@article{ma2020rose,
  title={{ROSE}: a retinal {OCT}-angiography vessel segmentation dataset and new model},
  author={Ma, Yuhui and Hao, Huaying and Xie, Jianyang and Fu, Huazhu and Zhang, Jiong and Yang, Jianlong and Wang, Zhen and Liu, Jiang and Zheng, Yalin and Zhao, Yitian},
  journal={IEEE Transactions on Medical Imaging},
  volume={40},
  number={3},
  pages={928--939},
  year={2020},
  publisher={IEEE}
}

@inproceedings{ronneberger2015u,
  title={U-net: Convolutional networks for biomedical image segmentation},
  author={Ronneberger, Olaf and Fischer, Philipp and Brox, Thomas},
  booktitle={International Conference on Medical Image Computing and Computer-Assisted Intervention},
  pages={234--241},
  year={2015},
  organization={Springer}
}

\end{document}